\documentclass[lettersize,journal]{IEEEtran}
\usepackage{amsmath,amsfonts}
\usepackage{algorithmic}
\usepackage{algorithm}
\usepackage{array}
\usepackage[caption=false,font=normalsize,labelfont=sf,textfont=sf]{subfig}
\usepackage{textcomp}
\usepackage{stfloats}
\usepackage{url}
\usepackage{verbatim}
\usepackage{graphicx}
\usepackage[justification=centering]{caption}
\usepackage{cite}
\begin{document}

\title{New Approach for an Affective Computing-Driven Quality of Experience (QoE) Prediction}

\author{
\IEEEauthorblockN{Joshua Bègue\IEEEauthorrefmark{1},
Mohamed Aymen LABIOD\IEEEauthorrefmark{1},
Abdelhamid MELLOUK\IEEEauthorrefmark{1}}

\IEEEauthorblockA{\IEEEauthorrefmark{1} University of Paris-Est Creteil, LISSI, TincNET (CIR), F-94400, Vitry-sur-Seine, France\\
Email: joshua.begue@u-pec.fr, mohamed-aymen.labiod@u-pec.fr, mellouk@u-pec.fr}

\thanks{This paper was produced by the IEEE Publication Technology Group. They are in Piscataway, NJ.}
\thanks{Manuscript received December 01, 2022.}}

\markboth{Journal of \LaTeX\ Class Files,~Vol.~14, No.~8, August~2021}%
{Shell \MakeLowercase{\textit{et al.}}: A Sample Article Using IEEEtran.cls for IEEE Journals}


\maketitle

\begin{abstract}

In human interactions, emotion recognition is crucial. For this reason, the topic of computer-vision approaches for automatic emotion recognition is currently being extensively researched. Processing multi-channel electroencephalogram (EEG) information is one of the most researched methods for automatic emotion recognition. This paper presents a new model for an affective computing-driven Quality of Experience (QoE) prediction. In order to validate the proposed model, a publicly available dataset is used. The dataset contains EEG, ECG, and respiratory data and is focused on a multimedia QoE assessment context. 
The EEG data are retained on which the differential entropy and the power spectral density are calculated with an observation window of three seconds. These two features were extracted to train several deep-learning models to investigate the possibility of predicting QoE with five different factors. The performance of these models is compared, and the best model is optimized to improve the results. The best results were obtained with an LSTM-based model, presenting an F1-score from 68\% to 78\%.
An analysis of the model and its features shows that the Delta frequency band is the least necessary, that two electrodes have a higher importance, and that two other electrodes have a very low impact on the model's performances. \\

\end{abstract}


\section*{Introduction}

Humans' responses to interactions in daily life are known as emotions. They may show themselves through a variety of human expression techniques, including psychophysiology, gesture, or biological responses. Analyzing emotions using computer vision and affective computing is seen as a fundamental challenge \cite{refEmotion}. However, in order to analyze emotions, it is necessary to detect the appropriate signals beforehand. 
Numerous methods exist to achieve this, including using non-physiological cues such as speech, body posture, or facial expressions. Physiological markers like heart rate, breathing, or brain signals like functional magnetic resonance imaging (fMRI) or electroencephalography (EEG) are also investigated to detect them \cite{refOVPD}.

On the other hand, in the network field, techniques and concepts are evolving rapidly, toward better performance, and lower latency, to keep up with the rapid increase in internet traffic. Among these, the emergence of the quality of experience (QoE) concept did not go unnoticed. It had an impact on almost every aspect of network environments, regardless of the application or context involving a network \cite{refBook}. The procedure to implement such QoE-aware network solution is already really well developed and can include many parts of the network equipment \cite{refQoEVideo}, creating a control loop whose goal is to maximize QoE.\\   
The QoE concept has truly changed how networks are designed. Indeed, before the use of QoE as the metric to measure the performance of any network application, Quality of Service (QoS) was at the center of most approaches with a QoS maximization control loop, which is always relevant and used, but QoE makes it possible to go further. The QoS is traditionally measured through the technical characteristics of the said application in the different fields concerned, such as the network and the video domains. It comes from the belief that better values upon these characteristics enable a better quality for the proposed service which is true in general. However, the opposite is not necessarily true; in fact, moderate impairments have been shown to have very little impact on users' perceived quality. \\
This is what the QoE is about; "How bad can the QoS become without having an impact on the user experience?" is a great way to summarize the idea of a QoE-driven approach to quality assessment. This is a significant change since maintaining a high QoS can be difficult and expensive.

Therefore, QoE can help to reduce end-to-end service complexity and the time used to develop a service for a similar product at the end (from the user's point of view). \\

Ultimately, knowing the user's feedback realistically will help to meet his expectations, making him more likely to pay for this service \cite{refBook}. From there, being able to do an accurate QoE estimation is very important. \\ 
Nevertheless, the QoE estimation oriented towards affective computing is still under-explored, with the use of physiological signals, such as EEG. \\

Because brain signals are the ideal place to investigate several cognitive mechanisms, the affective computing sector has taken off thanks to recent advancements in EEG systems. Since its advent, the domain has been expanding, with more and more work falling under its banner. \\
Part of the QoE prediction area focuses on trying to achieve a good prediction using psychophysiological signals such as an electrocardiograms and an electroencephalograms to obtain good prediction, rather than focusing on QoS factors. It is now well known that the QoE is influenced by other factors such as age, preferences in content, mood, and more. The QoE assessment is then a cognitive mechanism by itself, and its estimation is actually a part of affective computing.

In this article, we investigate a deep learning-based affective computing-driven approach to quality of experience assessment in a multimedia context. A processing pipeline presented here allows the extraction from raw EEG signals of the Differential Entropy (DE) and the Power Spectral Density (PSD) with an observation window of three seconds. The two features are calculated from five different frequency bands, permitting QoE estimation. Our estimation is performed through an LSTM-based deep learning model that has been chosen based on a comparison with two other models. After optimization, the selected model is achieving up to 78\% in F1-score, standing in state-of-the-art on this dataset.
The contributions of this work are as follows: 
\begin{itemize}
    \item We present a pipeline to process EEG data and extract features with which QoE estimation can be performed.
    \item Three Deep Learning-based architectures are proposed for QoE prediction on which a comparison took place to determine the most efficient model for this topic. 
    \item An evaluation of the importance of features and electrodes is performed, the results of which simplify and improve the procedure and the estimation.
\end{itemize}


\section*{Background and related work}
\label{related_work}

The QoE estimation field is now very large, with studies about the impact on QoE of network hardware, architectures, or even algorithms. \\
In the literature, QoS-based QoE prediction is widely used, and the research field is still active. In \cite{refQoEABR}, Duanmu et al. propose a publicly available dataset containing network and video features, generated by several algorithms used in streaming applications named adaptive bitrate (ABR), and propose an analysis of the said algorithms to enhance their QoE optimization. The solutions take into account the bandwidth available to exploit it at the maximum. Their findings indicate axes of enhancement, such as through a better understanding of the human visual system and psychological behaviors. \\
In \cite{refQoEBayes}, the authors proposed an objective Bayesian model to estimate a function representing the QoE based on knowledge of the human visual system, which they named the "Bayesian streaming model index" (BSMI). The proposed modeling has been optimized on existing datasets and tested on publicly available databases. The results presented are Pearson and Spearman correlations between the model prediction and the real mean opinion score (MOS). Among the existing objective models, the BSMI proposes the best results, with a correlation of 79.3\% on average over four datasets, compared to 76.4\% for the best overall model (on these same datasets). \\
Other works propose some modifications to the traditional estimation method. For example, Tiotsop et al. \cite{refQoERange} propose to estimate the QoE in intervals instead of the simple MOS value returned by the participant to give more room to the model while still maintaining the prediction relevance. In fact, the traditional approach is to try to predict the value directly given by the participant. Using well-known video quality metrics, their proposal gives an encouraging result, opening the way to that new type of prediction. \\
In the same way, the work investigating new features for their correlation with QoE is nonetheless important. In \cite{refAccAnn}, a new method for conducting QoE assessment experiments is presented, allowing for the management of acceptability and annoyance factors in a single step, thereby reducing the time required for these time-consuming QoE assessments. The proposed method is shown to lead to similar results as the traditional method while being way more convenient for the subjects. They show that users expectations can be modified by the instructions given to the subjects, and we know that these can have an impact on QoE, as psychological factors. \\
Because of the cognitive nature of QoE in general, the affective computing domain and its development have inspired many domains, including QoE estimation. In \cite{refQoEAffectiveGA}, Kitao et al. used affective computing factors and electroencephalographic data to estimate the QoE in a video streaming context. This work proposes a way to select EEG features using a generative algorithm (GA). Over 400 features were used, and they showed that the number of features necessary to predict QoE from EEG does not need to be very large. Using an SVM, they improved its performance up to 6\% in accuracy with their feature selection method, compared to a random feature selection. \\
The affective computing domain is filled with studies using electroencephalogram (EEG) data and associated analysis. For instance, in \cite{refMulti}, Rudakov et al. worked on the use of emotion estimation using EEG signals. Exploiting a famous publicly available dataset, the DEAP (Dataset for Emotion Analysis using Physiological signals) dataset, they propose a multi-task model, which is a very interesting approach as a single model is trained on several tasks, reducing the training time necessary. They achieved an accuracy of 96,28\% on valence and 96,62\% on arousal by training this model on PSD heatmaps of each frequency band of the EEG and differential entropy heatmaps. \\
The success of such work has already shown the potential of the EEG sensor, and many have since tried it in several domains. QoE and EEG have since been correlated, for instance in \cite{refEEGQoE}, where Kroupi et al. investigated the EEG signals for patterns associated with QoE. Among their findings, they found an asymmetry in the alpha band of the frontal electrodes, which appeared during low-quality perception. \\
Among the various works listed, some points proposed in this work have not yet been investigated, such as the EEG characteristics used here (in particular differential entropy) and their importance in the evaluation of QoE in a multimedia context. This is also valid for different deep learning models, which currently show the best results for the EEG data classification, especially those based on an LSTM model. These further investigations deserve greater attention to obtain better results and possibly simplify the process.

\section*{Proposal}
\label{Proposal}

In this paper, we are using a publicly available dataset, named SoPMD, which is made for the QoE estimation using psychophysiological data. This dataset has been used a few times, and all the data present has not been fully exploited. We aim to use the EEG data to predict the QoE using Machine Learning (ML) and especially deep learning (DL) techniques.

\begin{figure*}[!ht]
    \centering
    \includegraphics[width=\textwidth]{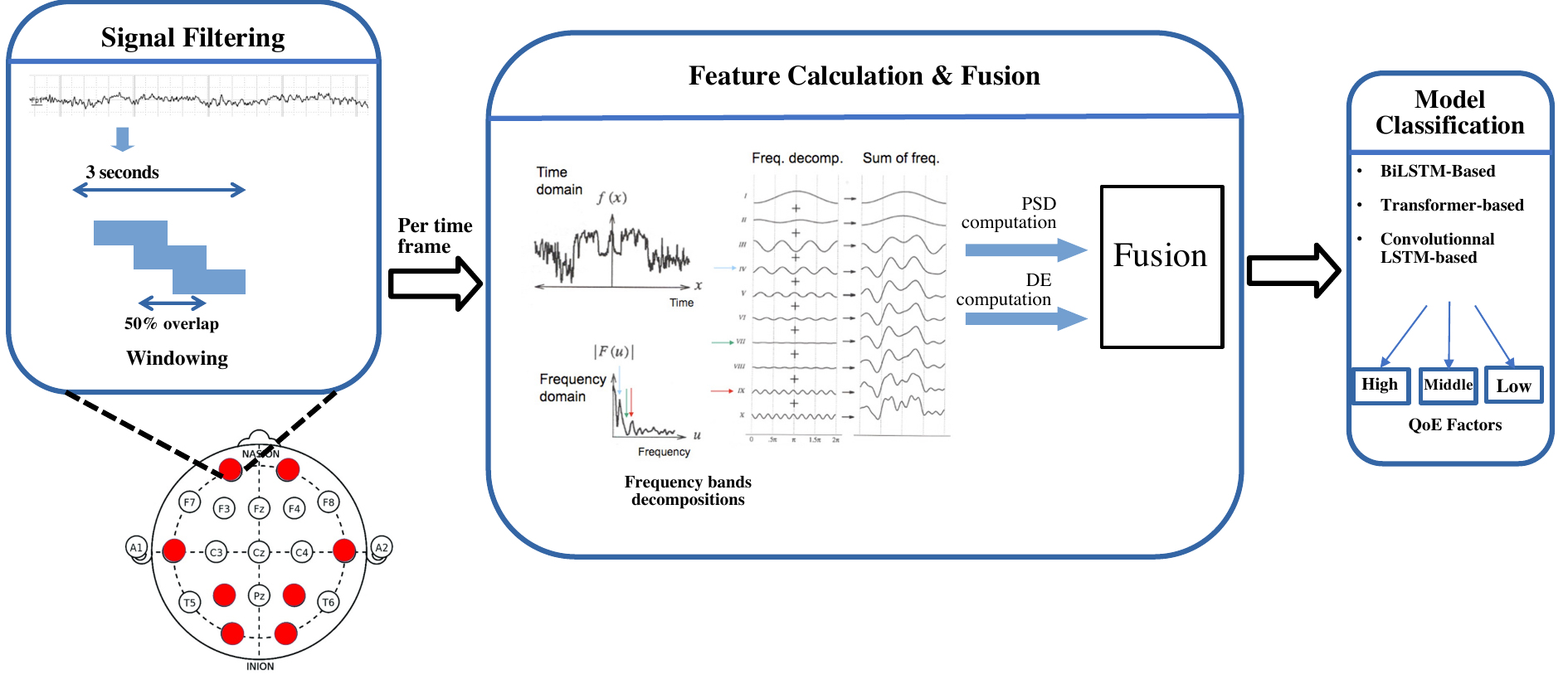}
    \caption{Data processing pipeline}
    \label{fig:pipeline}
\end{figure*}

\subsection*{Dataset}

The used SoPMD dataset has been introduced in \cite{refDataset}, where Perrin et al. created the dataset for QoE estimation from physiological activity. The experiment during which the data were collected was organized as the following. 
Before the experiments, subjects were required to wear an EEG headset, an ECG, and two respiratory bands. \\
The acquisition consists of three sessions, interspersed with breaks, in which subjects would watch nine 60-second long videos (each preceded by a 10-second baseline) and assess their experience by answering five questions measuring the QoE factors being: "immersive level" (IL), "perceived video quality" (VQ), "surrounding awareness" (SA), "interest in video content" (VC), and "interest in audio content" (AC) on a nine-point scale each. Subjects were not limited in time during the rating phase. \\ 
The chosen videos are from nine sequences, each being declined in terms of compression level (high and low QP, with x264 encoding), resolution (SD, HD, UHD), and audio sound system (mono, stereo, 5.1).\\
As required in \cite{refITU}, participants used a 56-inch Sony Trimaster SRM-L560 monitor to present stimuli and were seated at a distance of 1.6 times the height of the screen. Audio stimuli were played on an Altec Lansing 5.1 THX speaker system. The environment was quiet, and the ambient light were unobtrusive in both dark and bright scenes.
A total of 20 participants were involved to create the dataset. \\
The authors proposed a solution to estimate the immersion level. However, they have exploited all the signals: EEG, ECG, and respiratory, to extract characteristics to train an SVM model. A functional connectivity analysis was used to extract the EEG features. The model was able to achieve an average accuracy of 55.33\% on a three classes classification problem.\\
Vijayakumar et al. \cite{refBiLSTM} proposed also BiLSTM architecture to achieve the same prediction based on this dataset. In their proposal, they used 19 features from physiological data only (ECG and respiration signals), forming a total of 57 features to feed the model. With the help of a SMOTE data augmentation, their approach achieved an f1 score ranging from 58\% to 67\% depending on the evaluated QoE factor.

\subsection*{Psychophysiological data extraction}

\subsubsection*{Data preprocessing}

As previously mentioned, this dataset provides EEG, ECG, and respiration recordings from the subjects during their participation in the experiment. This paper presents a work using only the collected EEG data during this experiment, which is summarized in Fig. \ref{fig:pipeline}.\\
The EEG headset used for the experiment uses 256 electrodes to record brain activity at a frequency of 250Hz. We selected eight well-known electrodes for their ability to show the activity of each external brain region, to keep the spacial information. The electrodes used are as follows : Fp1 ; Fp2 ; T3 ; T4 ; P3 ; P4 ; O1 ; O2.\\
The raw signals cannot be used directly, EEG signals being very sensitive to noise and artifacts. To get rid of the noise, a bandpass filter with cut-off frequencies of 1 Hz and 47 Hz. The frequencies below 1 Hz are not important to us, and the high cut-off frequency was chosen to eliminate the 50 or 60 Hz perturbation due to electrical surroundings. Artifacts have not been removed as they are not very present.

\subsubsection*{Feature extraction}

In EEG analysis, the most used approach is the frequency analysis, to regroup the frequency energies or power in five bands, for their simultaneous appearance during several tasks or states (Delta band activity appearing during unconscious states, for instance) : 

\begin{itemize}
    \item $\delta$ (delta) : 1 to 4 Hz
    \item $\theta$ (theta) : 4 to 8 Hz
    \item $\alpha$ (alpha) : 8 to 13 Hz
    \item $\beta$ (beta) : 13 to 30 Hz
    \item $\gamma$ (gamma) : 30 to 60 Hz
\end{itemize}

Numerous features can be calculated from such spectral analysis, and among these, the most used and known is the power spectral density (PSD) from the five frequency bands introduced above. It was also important to include the other well-known type of feature, from the family of entropy-based features : the one selected is the differential entropy (DE) one. \\
The DE is a feature that has emerged in affective computing papers (but not only), as it was introduced by Shi et al.\cite{refDE} for the context of vigilance estimation. This feature is the measure the complexity of a continuous random variable, and the article from Shi et al. shows the relationship between their novel feature and the logarithm energy spectrum, widely used in EEG signal analysis. \\
The article is presenting how it can be computed, following the equation :
\begin{equation}
h_i (X) = \frac{1}{2}log(2\pi e\sigma^2_i)\
\label{DiffEntropy}
\end{equation}
Where $\sigma$ is the signal variance of the frequency band \textit{i}. It has been used in various contexts since.\\
To extract these features, it was chosen to use three seconds long segment as the window to perform the welch transform, with 50\% of overlap between windows, creating one data point for every 1.5 seconds. This gives us 40 points for each of the 10 features per acquisition, leaving a data shape of (40,80) as input for our model.

\subsection*{QoE classification method}

In affective computing, the most used tool for this part are DL classification models, and the results proposed are generally presenting outstanding results, if the models are well constructed regarding the addressed problem. \\
Inspired by literature, the results proposed by studies are significant enough for this approach to be an answer to the problem investigated here. More precisely, some models architectures have been selected for their potential in this context, each having already proved itself. Moreover, in EEG data processing and classification domain, most DL models are handling this data type very well and can outperform every other classification or regression technique known, in some issues. \\
The selected models architecture are the following :\\
\begin{itemize}
    \item LSTM-based models : The LSTM architecture is based on LSTM cells, standing for 'Long-Short Term Memory'. These models are a type of recurrent neural network (RNN) for their feedback connections, and the name is a reference its both 'short-term' and 'long-term' memory. The architecture we fancy with these is the possibility to use it in a bidirectional way, where data will go through two LSTM layers, with opposite direction of information (forward and backward). This type of model is often encountered in the classification or prediction of time-series data, as their 'memory' is an asset when using such data.\\
    
    \item Transformer-based models : The Transformer architecture has been introduced in 2017 with the goal of replacing LSTM architectures in natural language processing. For this model, the most important part is their self-attention mechanism, allowing them to gather information about a data point relative to its position (that can be named context) in a sequence. This is critical to know as, unlike RNN models, these process the entire input at once. These models are from the Encoder family.\\
    
    \item Convolutional LSTM-based model : This is also a type of model taking advantage of the LSTM cells strength, but also of Convolutional neural network (CNN) models. That allows these to collect local spatio-temporal information about the input data pretty well.
    
\end{itemize}

An implementation of each model has been done to produce preliminary results on the perceived video quality QoE factor, but only the one giving the best results at this time is going to be optimized at maximum for each factor. The one that was withheld after this selection was the BiLSTM-based, which was able to perform way better than the two other models tested, and was less subject to overfit (detailed below).

\subsection*{Model optimization}

For the models we selected, each layer can be adjusted by numerous parameters, named hyperparameters. For each different setting of each, we have a slightly different model, that can give various results. \\
In DL problems, this part is called model tuning, and is necessary if we want to have the model as optimized for our subject as possible. To address this issue, several methods exist and among which 'Gridsearch' is from. Its name come from the way the algorithm work, which is simple although resource-intensive: given a list of values for each hyperparameter and the model, the algorithm will simply train the model with every possible combination and keep track of the results for each training, viewable as a grid of values. At the end, the algorithm returns the parameters giving the best results, with the obtained results. \\
Applying such an algorithm has the advantage to result in the best model possible given a list of hyperparameters. The major method drawback is the time-consuming aspect since each value added to the list of parameters is adding exponentially more training. Given that model training can already be lengthy, depending on the complexity of the model and the data used with it, this technique exponentially increase the time required for the model hyperparameters adjustment.

\begin{figure*}[!ht]
    \centering
    \includegraphics[width=0.70\textwidth]{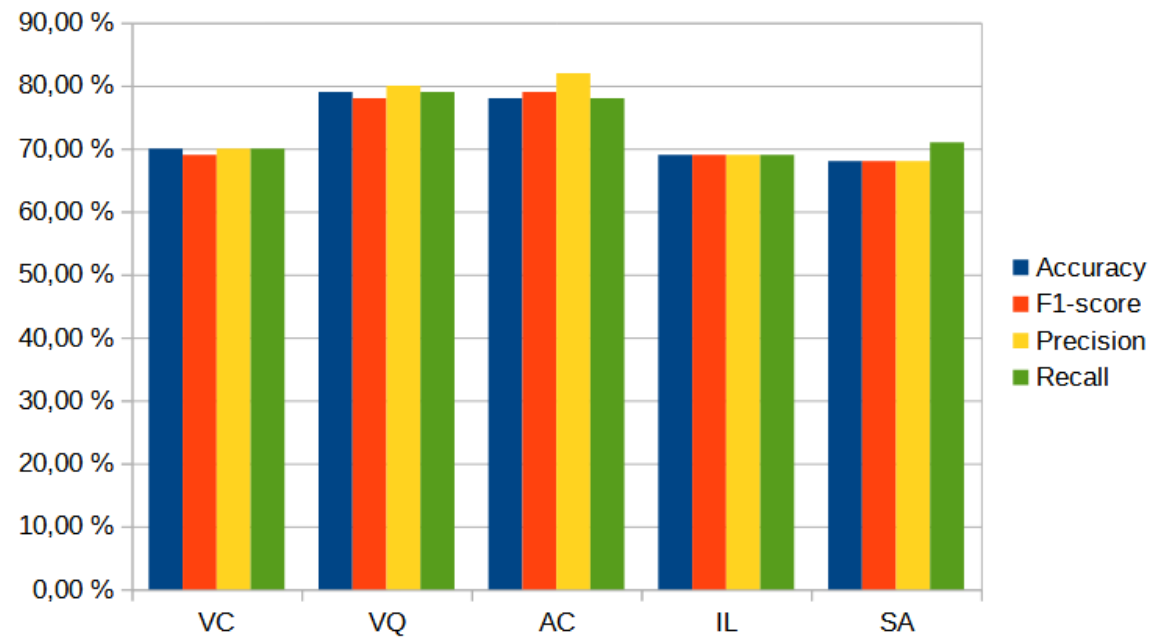}
    \caption{Results of the BiLSTM for each QoE factor}
    \label{fig:results_BiLSTM}
\end{figure*}

\section*{Results and discussion}
\label{results}

\subsection*{Classification results}

In this part, it is presented the results obtained with each model at the end of their training and after the processing of the entire pipeline mentioned before. \\ 

Before talking about the results, is presented first the parameters chosen to generate the said results. 
Data is separated into a training set and a testing set with 80\% of the dataset in the training part. The training was done with 100 epochs (or steps) as it was the best suited for our model from our tests, except for the Transformer, which had 150 epochs, and the best hyperparameters for each factor are presented in Table. \ref{hyperparameters} The model was trained on a three-class classification problem. Multiple mechanisms to prevent overfit were integrated such as :

\begin{itemize}
    \item Dropout layers, optimized for each factor with the GridSearch.
    \item L2 regularisation, also optimized for each factor with the GridSearch.
    \item Stratified K-fold, with 10 folds (unchanged during the GridSearch).
\end{itemize}

\begin{table}[!htbp]
\begin{tabular}{|l|l|l|l|l|l|}
\hline
                                 & VC  & VQ  & AC  & IL  & SA  \\ \hline
Bidirectional LSTM units layer 1 & 16  & 128 & 64  & 32  & 64  \\ \hline
Bidirectional LSTM units layer 2 & 128 & 16  & 128 & 64  & 128 \\ \hline
Dropout                          & 0,7 & 0,2 & 0,7 & 0,4 & 0,2 \\ \hline
L2 regulation                    & 0,2 & 0,6 & 0,4 & 0,6 & 0,2 \\ \hline
\end{tabular}
\caption{Values of hyperparameters for each factor after the Gridsearch optimization}
\label{hyperparameters}
\end{table}

As said in the classification part, the BiLSTM was the most efficient model, presenting results of 68\% to 79\% of accuracy for the five QoE factors, with the classes being low, middle, and high (for the concerned factor), as it is discussed below. \\

The model is constructed as follows:
\begin{itemize}
    \item Two bidirectional LSTM layers (with a tanh activation function, and an L2 kernel regularisation optimized with GridSearch), each followed by a BatchNormalisation and a Dropout layer (optimized with GridSearch).
    \item A single MLP block: two Dense layers (with respectively 128 and 3 units) separated by a Dropout layer (set to 0.3)
\end{itemize}

\begin{figure}[!h]
    \centering
    \includegraphics[width=0.45\textwidth]{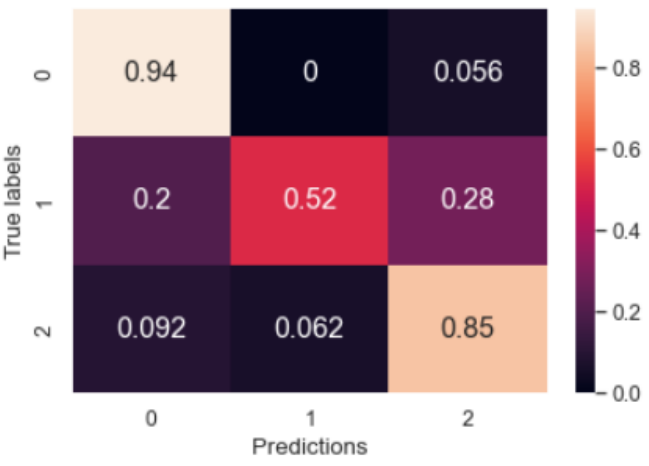}
    \caption{Confusion matrix of the test part for the perceived video quality QoE factor}
    \label{fig:cm}
\end{figure}

\begin{figure*}[!ht]
    \centering
    \includegraphics[width=1\textwidth]{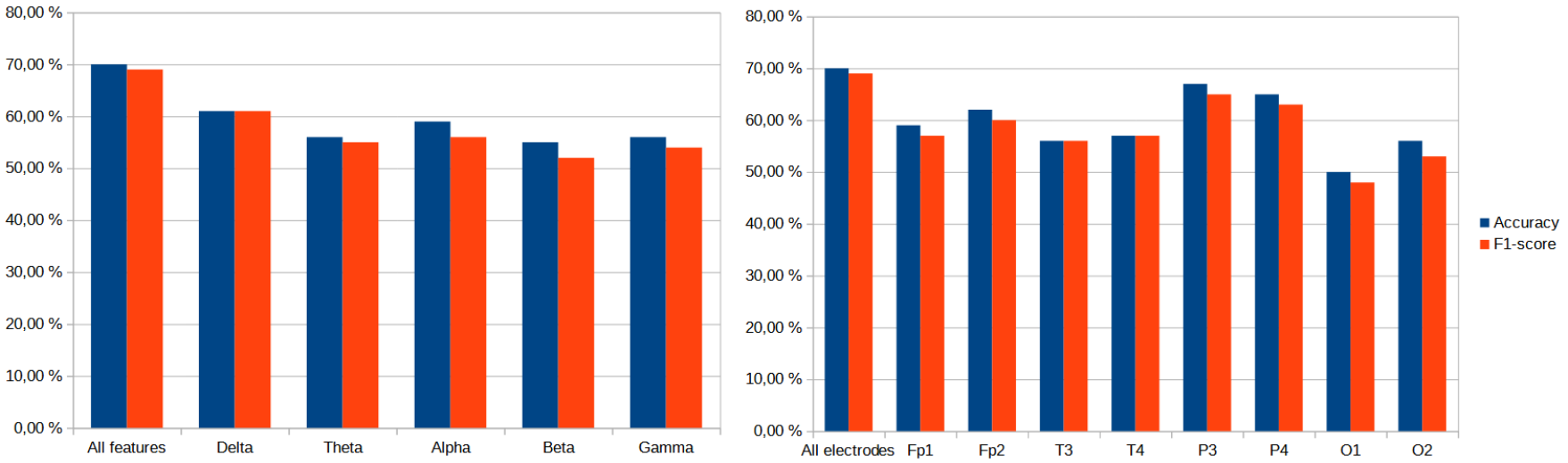}
    \caption{Results of the BiLSTM on the first QoE factor removing one frequency band (left) and removing one electrode (right)}
    \label{fig:Features_Electrode_imp}
\end{figure*}

In Table \ref{ModelSelection}, we present a part of the result which are the one for the perceived video quality for the three models, where the selected model was finally the best. In the table, we can see that the model was offering various results, placing overall the LSTM-based first with an accuracy of 79\% and an F1-score of 78\%, followed by the Transformer-based with 70\% of accuracy and an F1-score of 67\%, leaving the last place to the ConvLSTM-based with 61\% and 59\% for the accuracy and F1-score respectively.

In the Fig. \ref{fig:results_BiLSTM} are the four relevant performances indicators of the selected model for each factor, as a bar graph with: VC as interest in video content; VQ as perceived video quality; AC as interest in audio content; IL as immersive level; and SA as surrounding awareness. \\
It can be noticed that the model is presenting better results on the VQ and AC (78\% and 79\% of F1-score), and similar results on the VC, IL, and SA (with 69\%, 69\%, and 68\% of F1-score). As a reminder that the presented prediction accuracies are compared to the subjects' answers gathered during the experiments. A significant difference can be observed, of roughly 8\%. It can be explained by two reasons: it is either easier to evaluate VQ or AC in the brain signal, or it is easier to evaluate these for the subjects at the end of the experiments.

\begin{table}[!htbp]
\centering
\begin{tabular}{|l|l|l|l|l|}
\hline
Model       & Accuracy & F1-score & Precision & Recall   \\ \hline
BiLSTM      & 79,00 \% & 78,00 \% & 80,00 \%  & 79,00 \% \\ \hline
Transformer & 70,00 \% & 67,00 \% & 70,00 \%  & 70,00 \% \\ \hline
ConvLSTM    & 61,00 \% & 59,00 \% & 59,00 \%  & 61,00 \% \\ \hline
\end{tabular}
\caption{Results of the three models tested}
\label{ModelSelection}
\end{table}

In Fig. \ref{fig:cm}, a zoom on the results obtained for the perceived video quality is proposed, with the confusion matrix of the test part after the training phase. The prediction on the classes 'Low' (0) and 'High' (2)  are the best with 94\% and 85\% of accuracy respectively, which are near-perfect results. As expected, the class 'Middle' (1) is the hardest to classify for the model, since the boundary in the features from this class are not really defined, and even, 52\% of the examples from that class have been classified right. 

\subsection*{Feature importance} 

The knowledge of which feature, namely frequency band and electrode, has the most impact on the model is an important piece of information to us, for the future model optimization. In order to have an idea of this, it was decided to simply train the model while removing a frequency band, to see the impact it would have on the model prediction. The same method is used for the electrodes, and these results are presented in Fig. \ref{fig:Features_Electrode_imp}. For these two figures, the QoE factor on which the model is trained is the 'Interest in the video content (VC)', but any could have been selected.

First, for the frequency band importance, we see on the left of the Fig. \ref{fig:Features_Electrode_imp} that each frequency band has at least some necessary information for the estimator prediction, as removing any frequency band has an impact on the estimator results. If we look at this graph, it is obvious that removing the Delta band has less impact on the results, taking away 8\% from the original F1-score. This indicates that the Delta band contains the least information. It can be noticed that the other bands removal has a bigger impact than Delta, with Alpha, Theta, Gamma et Beta respectively taking away 13\%, 14\%, 15\%, and 17\% from the F1-score. From these results, we can classify their importance (from the more important to the less important) as: Beta, Gamma, Theta, Alpha, and Delta. \\
Secondly, the electrode importance can be seen on the right of Fig. \ref{fig:Features_Electrode_imp} and is presenting more differences. Two main things can be noticed : 

\begin{enumerate}
    \item P3 and P4 electrode removal have almost no impact on the results the model produce with 4\% and 6\% less on the F1-score. This shows that the parietal lobes are poor in information for the QoE estimation with the proposed model. 
    \item At the opposite, the removal of the O1 or O2 electrodes have the strongest impact on the results, with 21\% and 16\% lower F1-score.
\end{enumerate}

From the two importance analysis, it is clear that it could be possible to reduce the complexity of a real implementation while minimizing the impact on the results by removing the Delta frequency band from the feature extraction and the P3 and P4 electrodes.\\ 
The entire results presented in the article are encouraging for the possibility to read a user's QoE directly in the signals from his brain, to assess the quality of a multimedia application in a simpler and more efficient way, with state-of-the-art results. \\

\section*{Conclusion and future work}
\label{conclusion}
This paper presents an affective computing inspired approach for QoE estimation in multimedia applications. Using a publicly available dataset, features calculated from EEG and a BiLSTM-based deep learning model proposed an under-explored way to predict QoE, presenting results up to 79\% on accuracy and 78\% on F1-score. \\
We analyze the resulting model to show the importance of each EEG frequency band and electrode used. This will also contributes to future improvements and simplifications, for easier integration into existing systems. \\
These results could make it possible to improve the QoE evaluation of multimedia services by making it simpler and more efficient, particularly in virtual and augmented reality fields. In fact, these emerging next-generation multimedia technologies are currently only possible with the obligation to wear a dedicated helmet; this simplifies the addition of EEG electrodes in the helmet and would simplify the implementation of the proposed solution. Moreover, with a simplification of the needed data for the prediction, and improvement in the EEG field (in particular on the dry electrodes) an improvement of the results is strongly possible. 

As future work, we are investigating the use of all available data which should improve the current results. But the presented model needs to be adapted to be able to handle this, as it does not improve using all data for now. Overall, the results obtained in this article reveal the potential of affective computing techniques in the field of QoE prediction and it could still be improved with a complete description of the user's psychophysiological data. On the other hand, integrating this affective computing-based Quality of Experience (QoE) assessment into a QoE-maximizing control loop will be a big step forward. The real-time aspect of such a system must also be considered because the correction must be applied rapidly.

\section*{Acknowledgments}

We would like to thank the neurologist medical colleagues from the neurology department of the CHU Mondor in Créteil (France), in particular Dr. Yann Senova, for the many explanations they provided us.


\section*{Biographies}
\begin{IEEEbiography}[{\includegraphics[width=1in,height=1.25in,clip,keepaspectratio]{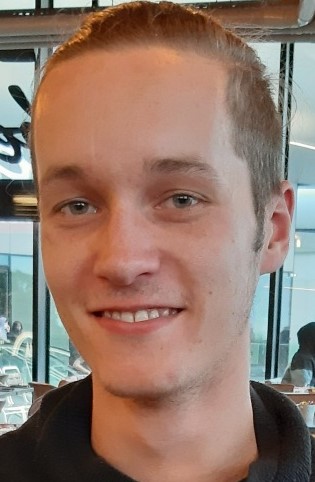}}]{Joshua Bègue}
PhD student at Paris-Est Créteil University (UPEC) and member of the Laboratoire Images, Signaux et Systèmes Intelligents (LiSSi) as part of the TINCNET research team. His PhD thesis focuses on the quality of experience (QoE) assessment using psychophysiological data and machine learning (ML) in a multimedia context.
\end{IEEEbiography}

\begin{IEEEbiography}[{\includegraphics[width=1in,height=1.25in,clip,keepaspectratio]{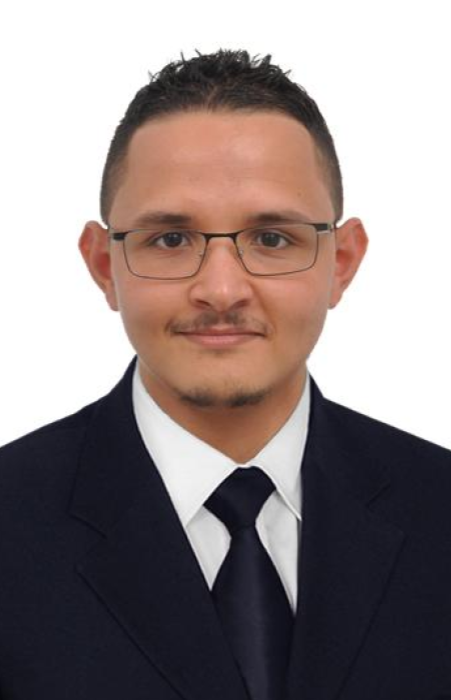}}]{Mohamed Aymen Labiod}
Associate professor at Paris-Est Créteil University (UPEC) and researcher at the Laboratoire Images, Signaux et Systèmes Intelligents (LiSSi) as part of the TINCNET research team. His main research themes focus are 5G and 6G architecture, network slicing, mobile networks, multi-access networks protocol, multi-connectivity, scheduling, adaptive resource allocation, cross-layer approaches, quality of experience (QoE) assessment, and multimedia communication under low latency constraints.
\end{IEEEbiography}

\begin{IEEEbiography}[{\includegraphics[width=1in,height=1.25in,clip,keepaspectratio]{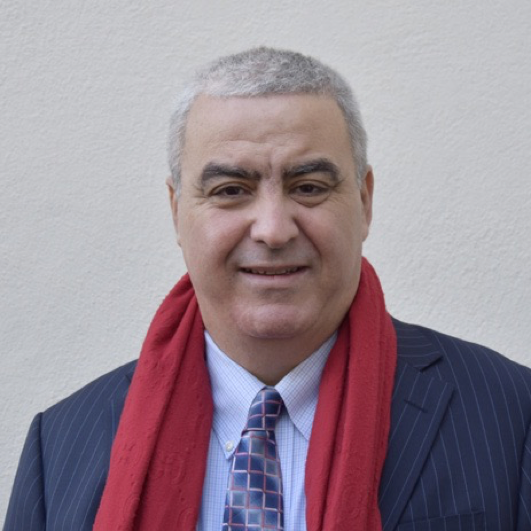}}]{Abdelhamid Mellouk} 

Full-time Professor at University of Paris-Est Creteil (UPEC), Networks Telecommunications (N\&T) Department-IUT CV, IT-Health Department-EPISEN and LiSSi-TincNET Research Team, France.  He is an active member of the IEEE Communications Society and held several offices including leadership positions in IEEE Communications Society Technical Committees. He has published/coordinated fourteen books and several refereed international publications in journals, conferences, and books, in addition to numerous keynotes and plenary talks in flagship venues. He serves on the Editorial Boards or as Associate Editor for several journals, and he is chairing or has chaired (or co-chaired) some of the top international conferences and symposia (including ICC and GlobeCom).
\end{IEEEbiography}

\vspace{11pt}


\vfill

\end{document}